\documentclass[letterpaper, 10 pt, conference]{ieeeconf}  

\IEEEoverridecommandlockouts                              

\usepackage{epsfig}
\usepackage{graphicx}
\usepackage{amsmath}
\usepackage{amssymb}
\usepackage{svg}
\usepackage{url}
\usepackage{adjustbox}
\usepackage{multirow}
\usepackage{multicol}
\usepackage{xcolor}
\usepackage{graphicx}
\graphicspath{{images/}}
\usepackage{setspace}

\title{\LARGE \bf
DESTINE: Dynamic Goal Queries with Temporal Transductive Alignment  for Trajectory Prediction
}

\author{Rezaul Karim$^{1}$, Soheil Mohamad Alizadeh Shabestary$^{2}$, and Amir Rasouli$^{2}$
\thanks{$^{1}$ EECS at York University. Work done while at Huawei. {\tt\small karimr31@yorku.ca}
}%
\thanks{$^{2}$Noah's Ark Laboratory, Huawei, Canada. {\tt\small soheil.shabestary@huawei.com}, {\tt\small amir.rasouli@huawei.com}}%
}

\begin{document}

\maketitle
\thispagestyle{empty}
\pagestyle{empty}

\begin{abstract}
      Predicting temporally consistent road users' trajectories in a multi-agent setting  is a challenging task due to unknown characteristics of  agents and their varying intentions. Besides using semantic map information and modeling interactions, it is important to build an effective mechanism capable of reasoning about behaviors at different levels of granularity.

To this end, we propose Dynamic goal quErieS with  temporal Transductive alIgNmEnt (DESTINE) method. Unlike past arts, our approach 1) dynamically predicts agents' goals irrespective of particular road structures, such as lanes, allowing the method to produce a more accurate estimation of destinations; 2) achieves map compliant predictions by generating future trajectories in a coarse-to-fine fashion, where the coarser predictions at a lower frame rate serve as intermediate goals; and 3) uses an attention module designed to temporally align predicted trajectories via masked attention.

Using the common Argoverse benchmark dataset, we show that our method achieves state-of-the-art performance on various metrics, and further investigate the contributions of proposed modules via comprehensive ablation studies.
\end{abstract}

\section{INTRODUCTION}
A key challenge in trajectory forecasting in the context of autonomous driving is modelling latent factors, such as  intentions of road users and their behavior while interacting with others. Existing approaches resort to explicit prediction of intentions in the form of multiple probable goals or destinations at the end of prediction horizon \cite{zhao2021tnt,gu2021densetnt, wang2022stepwise,lee2022muse}. Common techniques to achieve this goal include heuristic methods which rely on the scene layout or vehicles dynamics to estimate goals ~\cite{zhao2021tnt, chai2019multipath} and learning-based methods which rely on the data distribution ~\cite{lee2022muse, gu2021densetnt}. However, given the reliance on static context or past observations for goal prediction, these methods lack the ability to perform well in dynamically evolving cases where the intentions of the agents may change or the particular scenario has not been observed in the training data, i.e. out-of-distribution scenarios.  

To address the aforementioned shortcomings, a body of work focuses on designing dynamic architectures \cite{jia2016dynamic,yang2019condconv,zhou2021decoupled}, in which instead of directly estimating the model parameters, a series of filter kernels conditioned on the input are learned whose behavior adaptively changes during inference time allowing the model to adjust its behavior to current circumstances (see \cite{han2021dynamic} for more details). Inspired by similar approaches in the computer vision domain \cite{yang2019condconv,carion2020end,karim2023med}, we propose a dynamic goal predictor mudole that relies on a transformer-based architecture to generate dynamic goal queries during inference to estimate target locations. 

\begin{figure}
	\begin{center}
		\includegraphics[width=1\columnwidth]{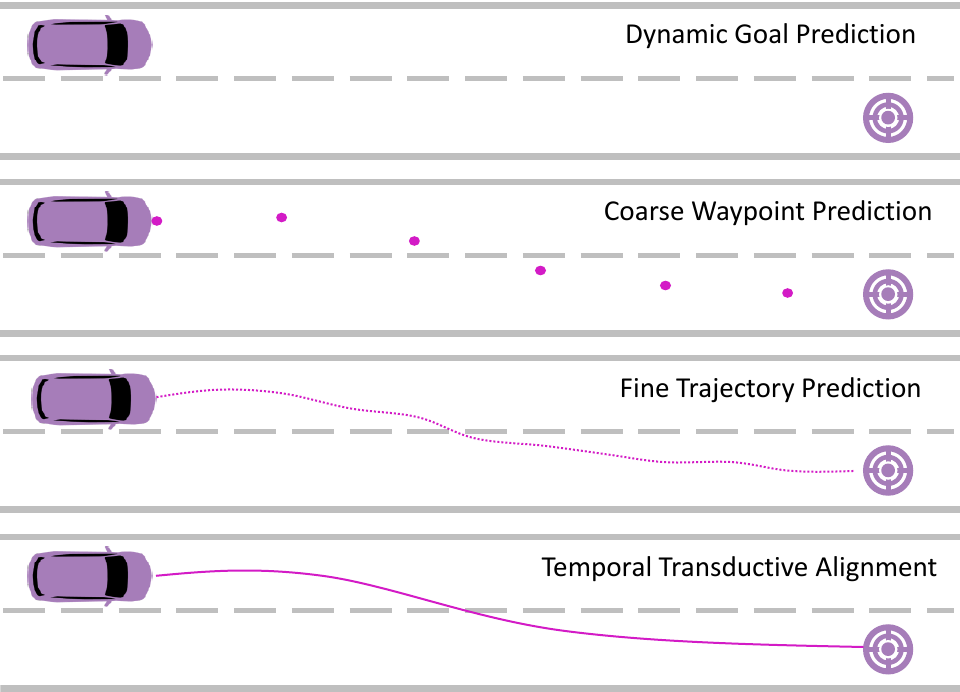}
	\end{center}
    \vspace*{-\baselineskip}
	\caption{An overview of the operation procedure of the propose model. Our method begins by dynamically predicting goal locations, followed by coarse waypoint prediction to define the intermediate points. Next, the model produces fine trajectories which are then fed into the temporal transductive alignment module for final refinement.}
	\vspace{-0.3cm}
	\label{fig:teaser}
\vspace*{-\baselineskip}
\end{figure}
Another existing challenge for prediction is to generate admissible trajectories, meaning they shoudl comply with road structure as well as dynamical constraints. To achieve addmissibility,  some approaches impose use autoregressive inference \cite{park2020diverse,amirloo2022latentformer}, heuristics \cite{Salzmann_2020_ECCV}, observation reconstruction \cite{Li_2021_ICCV, Yu_2021_IROS}, or rely on scene graph consistency computation \cite{Li_2022_CVPR} all of which can add significant computational overhead. To address this shortcoming, we propose a simple yet efficient temporal alignment mechanism that enforces consistency over the predicted trajectories based on cause-effect relationships.

In this paper, we introduce the Dynamic goal quErieS with  temporal Transductive alIgNmEnt (DESTINE) model to address the limitations of existing trajectory prediction models, as shown in Figure~\ref{fig:teaser}. Our model benefits from an attention-based architecture  that generates dynamic queries to estimate target locations as a proxy latent intents. Furthermore, our approach achieves road structure compliancy using a coarse-to-fine trajectory prediction scheme and a temporal transudctive alignment (TTA) mechanism. The coarse predictions, generated at a lower sampling rate, serve as intermediate goals to improve fine predictions while the TTA module aligns trajectory points across time. We conduct extensive empirical evaluations on the Argoverse \cite{chang2019argoverse} dataset and show that our method achieves state-of-the-art performance on various metrics. We further demonstrate the effectiveness of the proposed modules via ablation studies.

\section{Related Works}\label{sec:related}
\noindent\textbf{Intention Estimation}. There is a large body of work dedicated to trajectory prediction for autonomous driving, often specialized on pedestrians \cite{Rasouli_ICRA_2023, Rasouli_2021_ICCV, Shi_2021_CVPR, Li_2022_CVPR,Bae_2022_ECCV} and vehicles \cite{liang2020learning,kim2021lapred,zhou2022hivt,wang2022ltp}. Some of the main topics of interest in this domain include, scene representation \cite{cui2019multimodal,park2020diverse,gao2020vectornet,Casas_2020_ECCV, Cui_2021_ICCV, Ren_2021_ICCV,liu2021multimodal,ngiam2021scene,amirloo2022latentformer, Girgis_2022_ICLR}, interaction modelling   \cite{Helbing_1995_Phys,yamaguchi2011you,Rudenko_2018_ICRA,Kosaraju_2019_NeurIPS,Mohamed_2020_CVPR,Yu_2020_ECCV,park2020diverse,ngiam2021scene,yuan2021agentformer,Zheng_2021_ICCV,Xu_2022_CVPR_2,Da_2022_ICRA,  Girgis_2022_ICLR,zhou2022hivt,Hu_2022_ECCV}, latent intention modelling \cite{chai2019multipath,Salzmann_2020_ECCV,fang2020tpnet,zhao2021tnt,gu2021densetnt, zhang_2021_map,Kothari_2021_CVPR,yao2021bitrap,amirloo2022latentformer,lee2022muse,Varadarajan_2022_ICRA,xu2022remember,wang2022stepwise,lee2022muse}, and road structure compliance enhancement \cite{Lee_2017_CVPR,park2020diverse,Salzmann_2020_ECCV,park2020diverse,Li_2021_ICCV, Yu_2021_IROS,amirloo2022latentformer,Li_2022_CVPR}. 

One of the key challenges in trajectory prediction is modelling future uncertainty stemming from underlying agents' intentions which are not readily foreseeable. Some methods address this challenge by directly learning the distributions of trajectories over latent representations \cite{lee2022muse, yao2021bitrap, Salzmann_2020_ECCV} often conditioned on the future states of the agents in the training phase. These methods, however, are prone to mode collapse problem \cite{Casas_2020_ECCV, chai2019multipath} and can potentially become intractable as the space of possibilities grows. Alternatively, anchor-based approaches attempt to learn the space of possibilities, which is highly dependent on the quality of the hand-crafted anchors \cite{Varadarajan_2022_ICRA, Kothari_2021_CVPR, fang2020tpnet, chai2019multipath}. Heuristic based memory models estimates intentions from a memory database at inference time but their performance is shown to be limited when dealing with out-of-distribution scenarios \cite{xu2022remember}.

Another category of methods estimate, goals or potential target locations as a proxy to the intentions of agents \cite{wang2022stepwise,lee2022muse, zhao2021tnt,gu2021densetnt, zhang_2021_map}. Goals are predicted using cues, such as lane centerlines \cite{zhang_2021_map,zhao2021tnt}, area heatmap on map  \cite{lee2022muse}, or densely sampled drivable areas \cite{gu2021densetnt}. Given their reliance on static architectures, the effectiveness of these approaches is hindered when exposed to out-of-distribution scenarios. 

We address this issue by adopting a dynamic architecture for goal prediction inspired by \cite{jia2016dynamic,yang2019condconv,zhou2021decoupled,han2021dynamic}. Using a dynamic approach a subset of model parameters are adjusted during inference time allowing the model to adapt to scenarios that were not previously seen in the training phase. Given the success of this design appraoch in different applications of computer vision, such as classification~\cite{yang2019condconv}, object detection~\cite{carion2020end,dai2021dynamic}, and segmentation~\cite{fang2021instances,dong2021solq,karim2023med}, we an attention-based architecture that adapts model parameters to estimate targets using a dynamically learned goal queries  leading to better generalization to to out-of-distribution scenarios. 

\noindent \textbf{Compliant trajectory prediction.} To be reliable, generated trajectories should be compliant to (or consistent with) road structure and dynamical constraints. Existing models achieve this by observation reconstruction \cite{Li_2021_ICCV, Yu_2021_IROS}, scene graph consistency computation \cite{Li_2022_CVPR}, or heuristics methods \cite{Salzmann_2020_ECCV}. Alternatively, compliancy can be achieved by  autoregressive inference \cite{Lee_2017_CVPR,park2020diverse,amirloo2022latentformer}. However, besides the exposure bias problem \cite{schmidt2019generalization}, autoregressive processing is computationally expensive making it a less desirable choice for time-sensitive applications such as autonomous driving. Another line of work uses the coarse-to-fine approach in which intermediate goals (or waypoints) at a lower sampling rate are predicted as an intermediate steps to enforce compliancy \cite{mangalam2021goals, wang2022stepwise,lee2022muse}. We follow a similar scheme and additionally propose a  temporal transductive alignment (TTA) module which learns to align  generated trajectories using a masking operation. TTA is computationally efficient and operates on final generated trajectories allowing it to be applied to many existing prediction methods. \\

\noindent \textbf{Contributions} of our work are as follows:

\begin{itemize}
	\item We present a novel dynamic architecture for goal prediction that adaptively estimates road users' targets.
	\item We propose a novel temporal transductive model that aligns predicted trajectory points across time in a computationally efficient manner.
	\item We conduct extensive empirical evaluations on a public benchmark dataset and highlight the role of novel components using ablation studies. 
\end{itemize}

\section{Proposed Approach}
\subsection{Problem Formulation}
In our formulation, we use a continuous space discrete time sample assumption. Given a high dimensional map $M$ and observed states $S_{obs}=\{s^t_a: t \in T_{obs},  a \in A  \}$ of a multi-agent environment with agents set $A$($|A|=N$) for observation time steps  $T_{obs}= \{-t_o,...,0\}$, the task is to predict multimodal future trajectories, $S_{pred}=\{s^{t,k}_a|S_{obs}, M : t \in T_{pred}, a \in A,  k \in K \}$.  Here, $T_{pred}=\{1,...,t_p\}$ are future time steps, $K=\{1, ..., k\}$  are different modes of predicted trajectories, and $s^{t,k}_a$ is the future state of the agent. Each predicted trajectory is associated with a probability score $P=\{p^k_a: a \in A,  k \in K\}$ where $\sum_{k\in K}p^k_a=1$. 

\begin{figure*}[t!]
   \vspace*{+0.2cm}
	\begin{center}
		\includegraphics[width=0.99\textwidth]{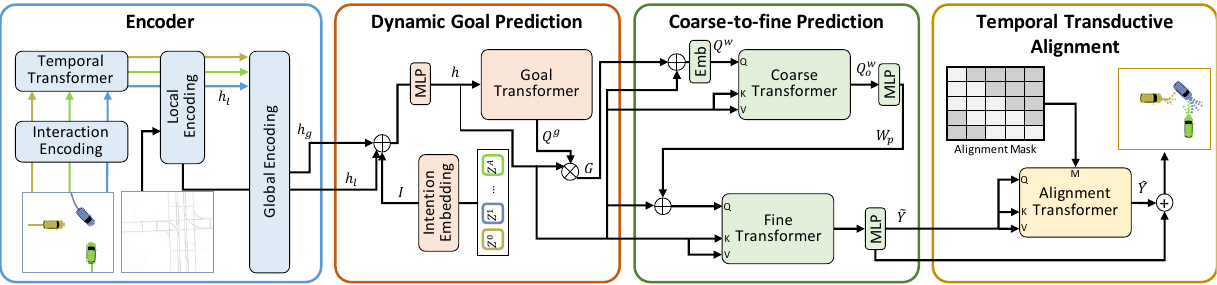}
	\end{center}
	\caption{An overview of the proposed approach. The encoder models the interactions between the agents and agents and the road. Next, a goal is generated that combined with encodings serves as the query to the coarse trajectory generator, the output of which is used to condition fine trajectory prediction. A the end, the TTA module temporally aligns the fine trajectories resulting in the final predictions.}
	\label{fig:model}
 \vspace*{-1.5\baselineskip}
\end{figure*}

\subsection{Model Overview}
An overview of our proposed method, DESTINE is depicted in Figure \ref{fig:model}. Overall, there are four core modules: \textbf{context encoding} where agents' dynamics along with high-dimensional map information are processed to produce a context representation; \textbf{dynamic goal prediction}, which receives the context encoding as input and learns goal queries to generate potential goal locations at time $t_p$ for the agents; \textbf{Coarse-to-fine trajectory prediction}, which relies on predicted goals and context encoding to generate trajectories; And \textbf{temporal transductive alignment} module refines the generated trajectories using a masked attention operation.

\subsection{Context Encoding}
The context encoding is inspired by the model in \cite{zhou2022hivt}, in which translation invariant represention of agents' attributes and the scene are generated by converting the coordinate reference frame to the center of agent at time $t=0$. The interactions between agents, and agents and the environment are modelled at two different levels. At the \textit{local} level, patches of radius, $r(=50m)$, centered at each agents are extracted. At each time step, the patches are processes using a self-attention layer to capture spatial relationships. The output of these patches are aggregated by concatenation and fed into a temporal transformer to model the temporal relations the output of which, in addition to lane information, are fed into an additional attention layer that captures the local lanes and agents relationships. This process produces a local agent $a$ representation $h_{a,l}$. At the \textit{global} level, a message passing operation in conjunction with applying a spatial attention is used to model the interaction between local representations. This produces a global agent representation $h_{a,g}$.

\subsection{Dynamic Goal Predictor}
We define goal as the final future position of predicted trajectory for each mode $k$, $s_a^{t_p,k}$. If estimated accurately, goals can help improve the compatibility of predictions to the road structure, i.e. result in admissible predictions that do not extend beyond the road boundaries~\cite{zhao2021tnt}. 

To generate the input to the goal prediction, we use an intention embedding layer based on a discrete set of mode representations, where a mode is a concrete instantiation of an intention, to enhance diversity and prevent mode collapse \cite{amirloo2022latentformer,Salzmann_2020_ECCV}. In particular, we use intention, $z^a$, for agent, $a$, with $K$ one-hot vectors of length $K$. This $Z=\{z^a:a \in A\}$ is then projected to $C$ dimensional embedding space by a linear layer,  added to a learnable positional encoding and finally concatenated with the index of agents to get intention embedding representation for all agents in the scene, $I$. Together, discrete intention embedding, $I=\{I_a:a\in A\}$, local features,$h_l=\{h_{a,l}:a\in A\}$, and global features $h_g=\{h_{a,g}:a\in A\}$ are concatenated to form the final feature encoding, $h$.

To enhance the generalizability of the goal predictor to out-of-distribution scenarios, inspired by dynamic architectures~\cite{jia2016dynamic,zhou2021decoupled,carion2020end,karim2023med}, the goal transformer is designed to generate dynamic filters in the form of goal queries to estimate a set of multimodal goals, $G$, from the observation encoding, $h$. We highlight the motivation behind this design, by comparing our learned dynamic goal queries to the non-dynamic learned goal predictor approaches similar to \cite{zhao2021tnt, gu2021densetnt} that use linear or convolutional filters to predict goals from the encoded observation features in Section \ref{sec:ablation}.

Traditionally, non-dynamic goal prediction models can be represented as $\mathsf{G} = \mathcal{F^s}( \mathsf{W^s}, \mathsf{h})$ where $\mathcal{F^s}$ is the goal predictor function , $\mathsf{W^s}$ is the model parameters, and $\mathsf{h}$  is the embedding from the encoder~\cite{zhao2021tnt,lee2022muse}. The parameter $\mathsf{W^s}$ is optimized during training from the entire training set and does not adapt during inference. Conversely, in the proposed dynamic query-based goal predictor, the operations are $\mathsf{Q^g} = \mathcal{F}^d(\mathsf{W^d}, \mathsf{h})$ and $\mathsf{G} = \mathcal{F}^q(\mathsf{Q^g},\mathsf{h})$. Here, $\mathsf{Q^g}$, $\mathsf{W^d}$, $\mathsf{h}$ and $\mathsf{G}$ correspond to goal queries, goal transformer parameters, input features and predicted goals, respectively. The function $\mathcal{F}^d$ is the goal transformer function that generates the goal query  $\mathsf{Q^g}$, while $\mathcal{F}^q$ is the goal prediction function. $\mathcal{F}^q$ is analogous to the operation of a linear layer in the multi-layer perceptron model which is a matrix multiplication with $\mathsf{Q^g}$ as a dynamic filer and  $\mathsf{h}$ as an input feature. In our model, goal transformer parameter  $\mathsf{W^d}$ is optimized during training to generate suitable goal queries $\mathsf{Q^g}$ to estimate goals, $\mathsf{G}$. Since the goal queries are analogous to dynamically generated weights or filters of a linear layer, the goal transformer takes a single input $h$ to be used as keys and values, while the queries are generated from a random initialization conditioned on input, $h$. In this context, we use the term \textit{dynamic} because the parameters, $\mathsf{Q^g}$, are generated at the inference time by the goal transformer. This formulation makes the goal predictor more adaptive to a given scene as the transformer learns how to focus on distinctive features generated by the network at the inference time \cite{arar2022learned}.

The goal transformer uses $l$ layers of multi-head attention with  the operation of  $i^{th}$ layer given by,
	\begin{equation}
     \label{eq:goal_decoder}
     \begin{gathered}
	\mathsf{Q}^g_i = 	\mathcal{M}_s(\mathsf{Q}^g_{i-1}+\mathsf{P}_q,\mathsf{Q}^g_{i-1}+\mathsf{P}_q,\mathsf{Q}^g_{i-1}) \\ 	
	\mathsf{Q}^g_i = 	\mathcal{M}_c(\mathsf{Q}^g_i+\mathsf{P}_q,h+\mathsf{P}_h,h) \\
	\mathsf{Q}^g_i = \mathcal{L}(\mathsf{Q}^g_i)
     \end{gathered}
	\end{equation}
\noindent where $\mathcal{M}_s$, $\mathcal{M}_c$, and $\mathcal{L}$ are self-attention, cross-attention~\cite{vaswani2017attention}, and a Multi Layer Perceptron (MLP) respectively. $\mathsf{Q}^g_i$, $\mathsf{P}_q$, $\mathsf{P}_h$ are the goal query at $i^{th}$ layer, query position embedding, and feature position embedding, respectively. $\mathsf{Q}^g_0$ begins with a random initialization and $\mathsf{Q}^g_l$ is the final goal query, $\mathsf{Q}^g$. Goals, $\mathsf{G}$ comprised of $B=5$ dimensional vector representing the mean and standard deviation of a Gaussian distribution of the 2D coordinates, and the confidence of the given mode. 

\subsection{Coarse-to-fine Trajectory Prediction}
We employ a multi-granular prediction scheme for improved map compliancy in which we initially predict coarse trajectories a lower sampling rate (e.g., 1 Hz) resulting in an output trajectory with sparser set of points. These points are often referred to as waypoints \cite{mangalam2021goals}, or intermediate or short term goals \cite{lee2022muse}. Given that for a specific start and end point an intermediate point in a short distance is unimodal \cite{zhao2021tnt}, we predict a single coarse trajectory for each goal. Our model predicts the intermediate points conditioned on both the goal and intention embeddings, and outputs predicted points as well as a new estimate of the goal. This allows the module to further refine the intially estimated goal. 

The coarse trajectory predictor relies on the transformer decoder formulation ~\cite{vaswani2017attention}, details of which is omitted for brevity. The predicted goal, $\mathsf{G}$, and the features encoding, $\mathsf{h}$, are concatenated to server as query input to the transformer decoder while  $\mathsf{h}$ serves as keys and values. Similar to the goal transformer, the inputs are processed using a self-attention, a cross-attention, and an MLP network to generate waypoints features followed by another MLP network to generate the coarse trajectory output $\mathsf{W}_p \in \{R^{N\times K\times T_{wp} \times B}\}$. Here, $T_{wp}$ is the number of waypoint samples. Next, the fine trajectory prediction module produces trajectories at the desired higher sampling rate (e.g., 10 Hz). The architecture of the fine predictor is similar to the coarse one where $\mathsf{h}$ and $\mathsf{W}_p$ are concatenated and used as the query input to the transformer decoder while $\mathsf{h}$ is used as keys and values. The fine predictor estimates final sampled trajectories $\mathsf{\tilde{Y}} \in \{R^{N\times K\times T_{pred} \times B}\}$. 

\subsection{Temporal Transductive Alignment}
Temporal transductive alignment (TTA) module is designed to enforce temporal cause-effect relationship over the time steps in non-autoregressive manner. Our design is motivated by the fact that real life driving trajectories are smooth over short time horizon resulted from the physical constraints of maneuvering vehicles. This means that the short-term trajectories should follow a consistent path without any apparent oscillating pattern. The proposed method re-aligns the predicted trajectories to achieve better consistency across different time-steps over a short time horizon, and consequently achieve better road structure compliancy.

TTA enforces temporal consistency within a short term temporal window by using a masked self-attention mechanism that is applied over the temporal dimension of the predicted trajectory, $\mathsf{\tilde{Y}}$, using an analytically designed mask $\mathsf{M_{t}} \in [0,1]^{T_{pred}\times T_{pred}}$, producing the final predictions $\hat{\mathsf{Y}}$. We define a temporal window $tw$ over which we want to enforce temporal consistency. In this process, the design of the attention mask is important. Through empirical evaluations, we identified a windowed square-subsequent masks to be the most effective in this context. The operation for a single layer of TTA can be summarized as follows, 
	\begin{equation} 
  \begin{gathered}
		\mathsf{Y^\prime} = \mathsf{\Tilde{Y}_{T\times (4\times N) \times K}}+\mathsf{P_{t}} \\
		\hspace*{-0.2cm} \mathcal{T}(\mathsf{Y^\prime}, \mathsf{M_{t}}) = \mathsf{S}\left(\frac{1}{\sqrt{d}}\mathsf{Y^\prime}\mathsf{W}^q_h(\mathsf{Y^\prime}\mathsf{W}^k_h)^\top+ \mathsf{M_{t}}\right)\mathsf{\Tilde{Y}}\mathsf{W}^v_h 
  \end{gathered}
     \label{eq:tta}
	\end{equation}
\noindent where $\mathsf{S}$ is the Softmax operation and $\mathsf{P_{t}}$ is learnable position embedding.  

  

\subsection{Learning Objective}
Our training loss is the combination of three objective functions given by, $L = \alpha L_g + \gamma L_{wp} + \beta L_{\hat{\mathsf{Y}}}$, where $L_g$, $L_{wp}$, and $L_{\hat{\mathsf{Y}}}$ are goal, waypoint (coarse predictions) and final trajectory losses respectively. Coefficients $\alpha$, $\gamma$, and $\beta$ are mixing weights which are set empirically. We compute the goal loss $L_g$, and as the negative log likelihood (NLL) and Huber loss with equal weights, and use only Huber loss for $L_{wp}$.  The final trajectory loss, $L_{\hat{\mathsf{Y}}}$, computes negative log likelihood (NLL), Huber loss with equal weights, and an additional classification loss using cross entropy for the confidence score. In all of the cases, we used only the best trajectory in terms of Final Displacement Errors (FDE) to compute the NLL and Huber loss.

\section{Experiments}
\subsection{Implementation Details}
The transformers used in our model all have 4 layers with 8 heads and embedding dimensions are set to 128. We chose 1Hz sampling rate for coarse prediction  and the standard 10Hz sampling for fine-grained predictions. We set the default masked window horizon $tw=2s$ in TTA. For training, We used teacher forcing scheme where ground truth end-points (goals) are used as input to the waypoint decoder for the first $60$ epochs and then we used the predicted goals to train the model for another $20$ epochs. The batch size was set to 32, initial learning rate $5e^{-4}$ with a cosine annealing learning rate scheduler~\cite{loshchilovsgdr}, and weight decay of $1e^{-4}$ using Adam optimizer \cite{loshchilovdecoupled}.
 
\subsection{Benchmarks}
\noindent\textbf{Dataset.} Existing datasets for vehicle trajectory prediction have significant diversity in data formats, annotation and evaluation protocols. For this reason it is a common practice to evaluate methods on a single large dataset~\cite{huang2022multi,liang2020learning,liu2021multimodal,zhou2022hivt} .

Following this approach and for a fair comparison with closely related works, we conduct our evaluation on the large scale Argoverse motion forecasting benchmark dataset~\cite{chang2019argoverse} on both validation and test subsets. \\

\noindent\textbf{Metrics.} As in the past arts \cite{Girgis_2022_ICLR,zhou2022hivt, ngiam2021scene}, we use the standard evaluation metrics, namely minimum Average Displacement Error ($minADE$), minimum Final Displacement Error ($minFDE$), and Miss Rate ($MR$).

Although distance-based metrics capture the accuracy  of predictions with respect to the ground truth, they do not highlight the admissibility or compliancy to road structure. For this purpose, we report on two additional metrics, namely Hard Off-Road (HOR) and Soft Off-Road (SOR) as proposed in \cite{Bahari_2022_CVPR}. The former calculates the percentage of predictions that at least have one predicted point of the trajectory off-road and the latter measures the percentage of all off-road points over all predicted points averaged over all scenarios. Unless otherwise mentioned, all evaluations are on $K=6$. For all metrics, lower values are better.\\

\noindent\textbf{Models.} We selected state-of-the-art models that have officially been evaluated on the Argoverse dataset. On validation set we report the results on TNT \cite{zhao2021tnt}, LaneGCN \cite{liang2020learning}, LaPred \cite{kim2021lapred}, MMTrans. \cite{huang2022multi},  AutoBot \cite{Girgis_2022_ICLR}, HiVT\cite{zhou2022hivt}, and  HLS-MM \cite{Choi_2022_ECCV}. To compute the new metrics, we used the official code released by the authors except for TNT\footnote{Implementation is from \url{https://github.com/Henry1iu/TNT-Trajectory-Prediction}}. For test set, in addition to the previous models that have test benchmark results, we also report on mmTransformer \cite{liu2021multimodal}, DenseTNT \cite{gu2021densetnt}, Scene Transformer \cite{ngiam2021scene}, and LTP \cite{wang2022ltp}.

\subsection{Comparison to State-of-the-art}
\begin{table}[]
\vspace{+0.2cm}
\caption{Evaluation results on the  Argoverse validation set for K=6. For all metrics lower values are better.}
\resizebox{\columnwidth}{!}{%
\begin{tabular}{l|c|ccccc}
\hline
\multicolumn{1}{c|}{Models} & Venue            & minADE            & minFDE              & MR                & HOR      & SOR \\ \hline
TNT\cite{zhao2021tnt}               & CoRL'20  & 0.95               & 1.73               & 0.21               & 3.00                & 0.15 \\
LaneGCN\cite{liang2020learning}     & ECCV'20  & 0.71               & 1.08               & 0.10               & 2.55                & 0.12 \\
LaPred\cite{kim2021lapred}          & CVPR'21  & 0.71               &1.44                &  -                 &   -                 &   -   \\
MMTrans.\cite{huang2022multi}       & ICRA'22  & 0.71               & 1.08               & 0.10               & 3.02                & 0.15 \\
AutoBot\cite{Girgis_2022_ICLR}     & ICLR'22  & 0.73               & 1.10               & 0.12               & -                   &  -  \\
HiVT\cite{zhou2022hivt}             & CVPR'22  & 0.66               & 0.97   & 0.09                           & 2.36    & 0.11 \\
HLS-MM\cite{Choi_2022_ECCV}         & ECCV'22  & 0.65   & 1.24               &-                                  &-                     &   -                \\
\hline
\multicolumn{2}{l|}{\textbf{DESTINE (Ours)}}                      & \textbf{0.64}    & \textbf{0.90}     & \textbf{0.08}             & \textbf{2.05}       & \textbf{0.10} \\ \hline
\end{tabular}}
\label{tbl:sota-val}
\vspace*{-\baselineskip}
\end{table}

\begin{table}[]
\caption{Evaluation results on the Argoverse test set, for K=1 and 6. For all metrics smaller value is better.}
\vspace{+0.2cm}
\resizebox{\columnwidth}{!}{%
\begin{tabular}{l|c|ccl|ccl}
\hline
\multicolumn{1}{c|}{\multirow{2}{*}{Models}} &
  \multirow{2}{*}{Venue} &
  \multicolumn{3}{c|}{K=1} &
  \multicolumn{3}{c}{K=6} \\ \cline{3-8} 
\multicolumn{1}{c|}{} &
   &
  minADE &
  minFDE &
  \multicolumn{1}{c|}{MR} &
  minADE &
  minFDE &
  \multicolumn{1}{c}{MR} \\ \hline
TNT\cite{zhao2021tnt}                 & CoRL'20  & 2.17 &4.96  &0.71  &0.98           &1.45  &0.17\\
LaneGCN\cite{liang2020learning}       & ECCV'20  & 1.70 & 3.76 &0.59  & 0.87          & 1.36 & 0.16 \\
mmTrans.\cite{liu2021multimodal}      & CVPR'21  & 1.74 & 4.00 & 0.60& 0.84          & 1.34 &  0.15  \\
DenseTNT\cite{gu2021densetnt}         & ICCV'21  & 1.68 & 3.63 & 0.58 & 0.88          & 1.28 & 0.13 \\
MMTrans.\cite{huang2022multi}         & ICRA'22  & 1.74 & 3.90 & 0.60  & 0.84          & 1.29 & 0.14  \\
Scene Trans.\cite{ngiam2021scene}     & ICLR'22  & 1.81 & 4.05 &0.59  & 0.80          & 1.23 & 0.13 \\
AutoBot\cite{Girgis_2022_ICLR}       & ICLR'22  & 1.84 & 4.12 & 0.63 & 0.89          & 1.41 & 0.16 \\
HiVT\cite{zhou2022hivt}               & CVPR'22  & 1.60 & 3.53 & 0.55 & \textbf{0.77} & 1.17 & 0.13 \\
LTP\cite{wang2022ltp}                 & CVPR'22  & 1.62 & 3.55 & 0.56 & 0.83          & 1.30 & 0.15 \\
\hline
\multicolumn{2}{l|}{\textbf{DESTINE (Ours)}}     &
  \textbf{1.59} &
  \textbf{3.49} &
  \textbf{0.54} &
  \textbf{0.77} &
  \textbf{1.15} &
  \textbf{0.12} \\ \hline
\end{tabular}%
}
\label{tbl:sota-test}
\vspace*{-\baselineskip}
\end{table}

We begin our experiments by evaluating the proposed method, DESTINE, against state-of-the-art trajectory prediction algorithms. The results of the experiments on the validation set are shown in Table \ref{tbl:sota-val}, in which we can see that our approach achieves the best performance on all metrics. On distance metrics, more notable improvement is achieved on FDE (by 7\%) and MR (by 11\%) highlighting the importance of effective goal selection and refinement. More improvements are achieved on admissibility metrics (by up to 13\% on HOR), showing that the generated trajectory samples by our model are generally more compliant with the road structure. A similar pattern of improvements is achieved on test set as shown in Table \ref{tbl:sota-test}. Once again the benefit of the proposed method is more apparent on FDE and MR metrics where improvement of up to 8\% is achieved. The comparison between 1 and 6 mode results shows that our model is not only better at generating good samples but also at selecting the most confident one. In terms of accuracy efficiency trade-off, we observe $6\%$ improvement in FDE at the cost of 25ms increase in runtime without any optimization (94ms vs 69ms) compared to HiVT~\cite{zhou2022hivt}. This current limitation of the computational overhead is worth the accuracy gain for safety critical applications, such as autonomous driving. 

\subsection{Qualitative Results}
\begin{figure*}
   \vspace*{+0.2cm}
	\begin{center}
		\includegraphics[width=0.95\textwidth]{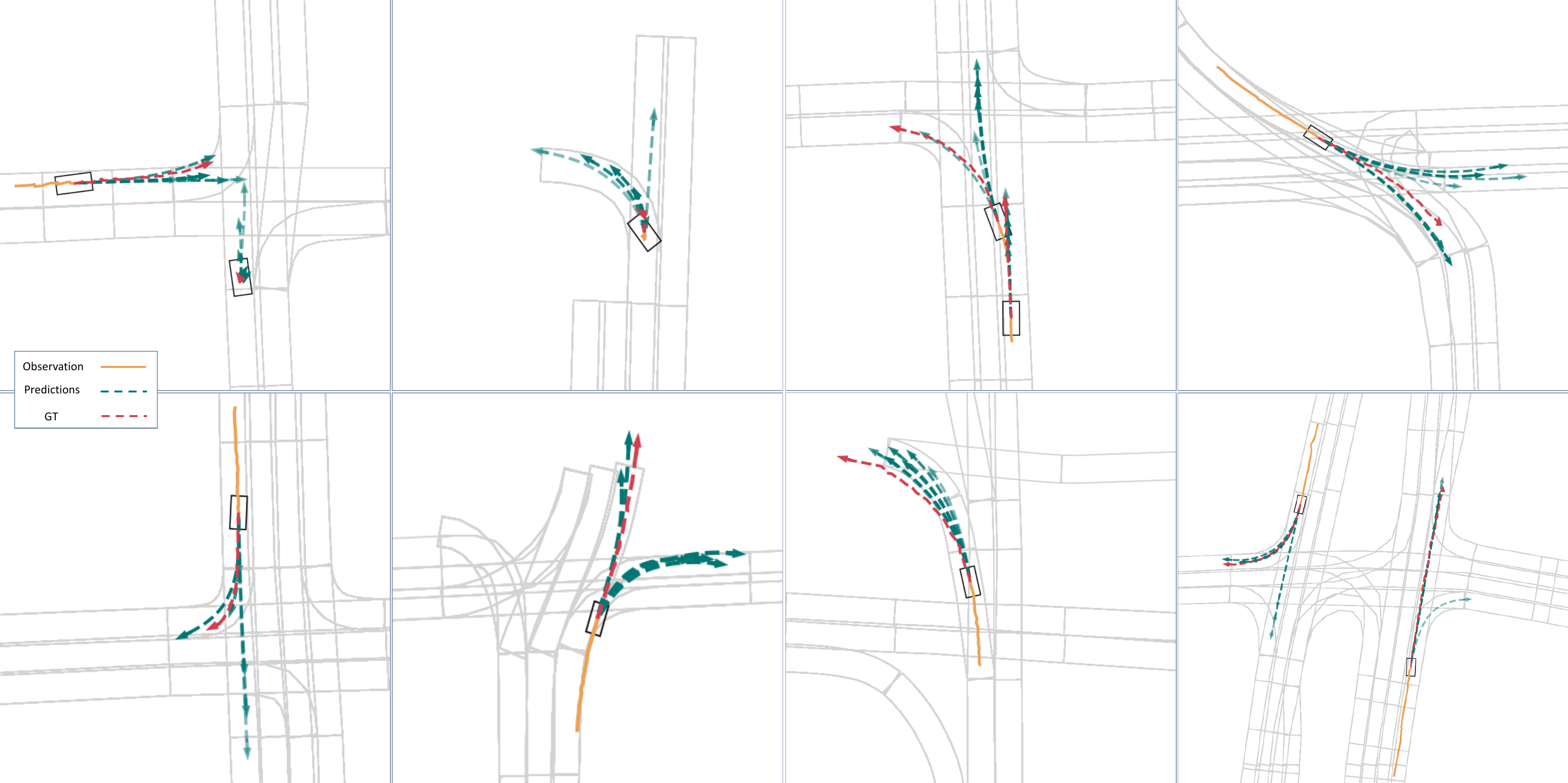}
	\end{center}\vspace*{-0.3cm}
	\caption{Qualitative examples of our proposed method, DESTINE, on the Argoverse validation set. The observation is shown in \textcolor{orange}{orange}, predictions are in shades of \textcolor{teal}{teal} with opacity as confidence, and ground-truth is in \textcolor{red}{red}.}
	\label{fig:qual}
\vspace*{-0.5cm}
\end{figure*}

Examples of our model's performance are depicted in Figure \ref{fig:qual} showing the effectiveness of DESTINE in capturing different modes of driving. Of particular interest is the scenario in which the vehicle is stationary (top-left sample) during observation phase. In this example, our method correctly deduces that the vehicle won't move forward by generating high confidence samples  while considering the possibility that the vehicle might move forward as captured by two modes of prediction. When the true intention of the vehicle is unclear (top-right sample), our model produces samples with high confidence for both possibilities ahead of the vehicle. 

\subsection{Ablation Analysis}\label{sec:ablation}
For ablation studies we report the results in terms of the displacement errors at some intermediate time steps (minDE@\#s) and MR of the best of the $K=6$ trajectories based on FDE. 

\paragraph{Contribution of Modules} We begin our ablation study by measuring the contribution of individual modules on the overall performance. The results are shown in Table \ref{tab:abl_modular}. Our base model is a simple \textit{trajectory} predictor that  only uses the fine predictor module, and we gradually add \textit{goal} predictor, \textit{waypoints} (coarse trajectory predictor), and \textit{TTA} module. As expected, adding goals provides a constraint on the end point of the prediction, hence improving on both distance metrics and MR. Using additional waypoints further improves the performance, in particular on MR. The application of TTA at the end has a significant impact, especially in terms of minimizing the error propagation towards the end point of the predicted trajectories, as a result, it further lowers the error on most metrics. Overall, the improvements become more prominent with the increase of time implying that simpler models are competitive for short futures but reliable long term prediction needs to utilize the benefits of the added modules.   

\begin{table}[t]
\vspace{+0.2cm}
    \caption{Ablation studies on the components of our model.}
	\begin{center}
		\begin{adjustbox}{max width=0.48\textwidth}	
			\begin{tabular}{l|c|c|c|c|c|c|c}
				\hline
				Trajectory  & Goal & Waypoints  & TTA & minDE@1s &minDE@2s &minDE@3s & MR \\
				\hline 
				\checkmark & - & -  & - &0.53& 0.84 & 0.96 & 0.090\\  
				\checkmark & \checkmark  & - &-  & 0.53& 0.81 & 0.94& 0.088  \\
				\checkmark & \checkmark & \checkmark & - & 0.52& 0.79& 0.93 & 0.084 \\
				\checkmark & \checkmark & \checkmark & \checkmark & \textbf{0.52}& \textbf{0.77 }&\textbf{0.90} & \textbf{0.081}\\		
            \hline
			\end{tabular}
		\end{adjustbox}
	\end{center}
	\label{tab:abl_modular}
\vspace*{-2\baselineskip}
\end{table}

\paragraph{Dynamic Goal Queries}
In this section, we highlight the advantage of the proposed dynamic goal query module. We implement an alternative goal predictor, which we term \textit{Learned Goal Predictor} that uses non-dynamic filter parameters learned during training time, similar to \cite{zhao2021tnt, gu2021densetnt}. In particular, we use 3 variants of \textit{Learned Goal Predictor}, using an MLP network  with  2, 4, and 6 layer. We compare these versions with our method, \textit{Dynamic Goal Predictor}, with 2 and 4 transformer layers. Both \textit{Learned Goal Predictor} and \textit{Dynamic Goal Predictor} use the same input, and they have similar structures for their outputs. Table \ref{tab:abl_dqd} summarizes the results of this study and shows the significant improvements of \textit{Dynamic Queries} over \textit{Learned Goal} decoder. We observe that the improvements on predictions are more significant towards the longer horizons, up to 7\% on MR, pointing to the importance of goal prediction to minimize long-term error propagation. Unlike the learned goals method, dynamic queries performance boosts further by adding extra processing layer showing the capacity of the method for learning better goal predictions. Overall, we see that dynamic goal queries is a better solution for reliable prediction in longer term horizons.  
\begin{table}[t]
	\caption{Comparison of alternative goal predictor architectures.}
	\begin{center}
		\begin{adjustbox}{max width=0.48\textwidth}	
			\begin{tabular}{l|c|c|c|c|c}
				\hline
				Goal Decoder & layers & minDE@1s&minDE@2s&minDE@3s&MR\\
				\hline 
				Learned Goal Predictor & 2 &  0.53& 0.79& 0.95 &0.088\\
				Learned Goal Predictor & 4 & 0.53 & 0.79& 0.94 &0.088\\
				Learned Goal Predictor & 6 &  0.53& 0.79& 0.94  &0.087\\
				\hline
				Dynamic Goal Predictor & 2 &  \textbf{0.52}& 0.78& 0.93 &0.083\\				
				Dynamic Goal Predictor & 4 & \textbf{0.52}& \textbf{0.77}& \textbf{0.90} &\textbf{0.081}\\
				\hline
			\end{tabular}	
		\end{adjustbox}
	\end{center}
	\label{tab:abl_dqd} \vspace{-0.3cm}
\end{table}

\paragraph{Temporal Transductive Alignment (TTA) }
\begin{table}[t]
 	\caption{Evaluation of different masking operations in  TTA module.}
	\begin{center}
		\begin{adjustbox}{max width=0.48\textwidth}	
			\begin{tabular}{l|c|c|c|c}
				\hline
				TTA & minDE@1s&minDE@2s&minDE@3s&MR\\
				\hline 
				- & 0.52 & 0.79 & 0.93 & 0.084  \\
				\hline
				1 layer/no-mask &  0.53 & 0.81 & 0.97& 0.094 \\
                1 layer/masked w history &  0.52 & 0.79 & 0.92& 0.086 \\
				\textbf{1 layer/masked} & \textbf{0.52} & \textbf{0.77} & \textbf{0.90}& \textbf{ 0.081}\\				
				2 layer/masked & 0.52 & 0.78 & 0.91& 0.081\\
				\hline
			\end{tabular}	
		\end{adjustbox}
	\end{center}
	\label{tab:abl_ttp}
\end{table}

To further probe into the contribution of TTA, we investigate on different variants of TTA with \textit{no-mask}, \textit{mask with history} in which observation is also included in consistency refinement similar to \cite{Li_2021_ICCV}, and our proposed approach, \textit{masked} with 1 and 2 layers of processing. As shown in Table \ref{tab:abl_ttp}, without masking information, the use of TTA can have a negative effect and degrades the performance. This can be due to the fact that the additional transformer operation adds noise to the prediction as the model tries to establish connections across all time steps. Adding observation to the masking operation, no noticeable improvement is achieved, however, focusing only on predicted trajectories  performance is boosted across all time steps, in particular at the longest prediction horizon, 3s. This behavior is expected since TTA minimizes error propagation across time resulting in less deviation from the ground truth trajectory. Adding extra layer of processing, no improvements was achieved.

\section{Conclusion}

In this paper, we proposed a novel trajectory prediction model, DESTINE,  comprised of dynamic goal prediction, coarse-to-fine prediction, and temporal transductive alignment modules. The proposed goal prediction module uses dynamic queries to adaptively predict goals at the inference time. The coarse-to-fine predictor provides intermediate waypoints serving as a condition for generating more stable trajectories. Lastly, the temporal alignment module refines the predicted trajectories  minimizing error propagation in long-term prediction. Via empirical evaluations on different sets of the Argoverse benchmark dataset, we showed that our model achieves state-of-the-art performance on various metrics, and further we highlighted the contributions of different proposed modules on the overall performance using ablation studies.

\section{Supplementary Materials}

\subsection{Dataset}
Each scene in the Argoverse dataset contains multiple vehicles sharing the road and out of all vehicles only one vehicle is specified for trajectory prediction. The dataset provides 333K 5-second long sequences of real world driving data with both trajectory history and high definition semantic maps. The trajectories are sampled at 10Hz, with first 2 seconds for observation and 3 seconds for future prediction. 
The data is divided into training, validation, and test sets containing 211K, 41K, and 80K trajectory samples respectively. The test set only has the first 2 seconds publicly available. For our experiments, we use both validation and test sets for comparison to past arts and use validation for our ablation studies.

\subsection{Ablation Study: Coarse Prediction}
\begin{table}[h]
	\begin{center}
		\begin{adjustbox}{max width=0.48\textwidth}	
			\begin{tabular}{l|c|c|c|c|c}
				\hline
				S. Rate (Hz) & minDE@1s&minDE@2s&minDE@3s& minADE &MR\\
				\hline 
				0 &  0.527 & 0.793  & 0.913  & 0.650  & 0.083    \\
                    1 &  \textbf{0.518} & \textbf{0.772}  & \textbf{0.904}  & \textbf{0.640}  & \textbf{0.081}    \\
				2 &  0.521 & 0.787  & 0.913  & 0.645  & 0.083    \\
				3 &  0.518 & 0.784  & 0.917  & 0.645  & 0.083    \\
				\hline
			\end{tabular}	
  \end{adjustbox}
	\caption{Comparison of the proposed module, DESTINE, without ($0$ rate) and with coarse prediction module using different sampling rates (S. Rate). For all metrics lower value is better and  the best results are shown as bold.}
 \end{center}
	\label{tab:abl_waypoints}
\end{table}

In this study, we examine the contribution of coarse (waypoint) predictor module on the overall performance. We report the results for coarse predictors with different sampling rates, 0-3 Hz,  where $0$ corresponds to no coarse predictor used. This study is insightful as the effect of waypoint generation frequency had not been investigated in the previous similar works \cite{mangalam2021goals, lee2022muse}. For all versions, only coarse predictor is changed while everything else remains the same as the original model, including 10Hz fine trajectory prediction. 

As shown in Table \ref{tab:abl_waypoints}, it is evident that using coarse (waypoints) prediction, at different sampling rates, can improve the performance across different metrics. However, the improvement gain varies at different rates. The most improvement across all metrics is achieved when waypoints are generated at 1Hz. Further increasing the rate results in decline in the performance. We speculate that increasing the number of waypoints make the output of the coarse predictor more similar to the fine predictor, and as a result, reduces the impact of its regularizing effect on the final trajectory predictions. In addition, from the results we can see that unlike the goal prediction that mainly impacts the end point predictions, coarse prediction can improve the error rate across different time-steps as highlighted in both minDE metrics as well as minADE.


\subsection{Qualitative Results}
\begin{figure}
	\begin{center}
		\includegraphics[width=1\columnwidth]{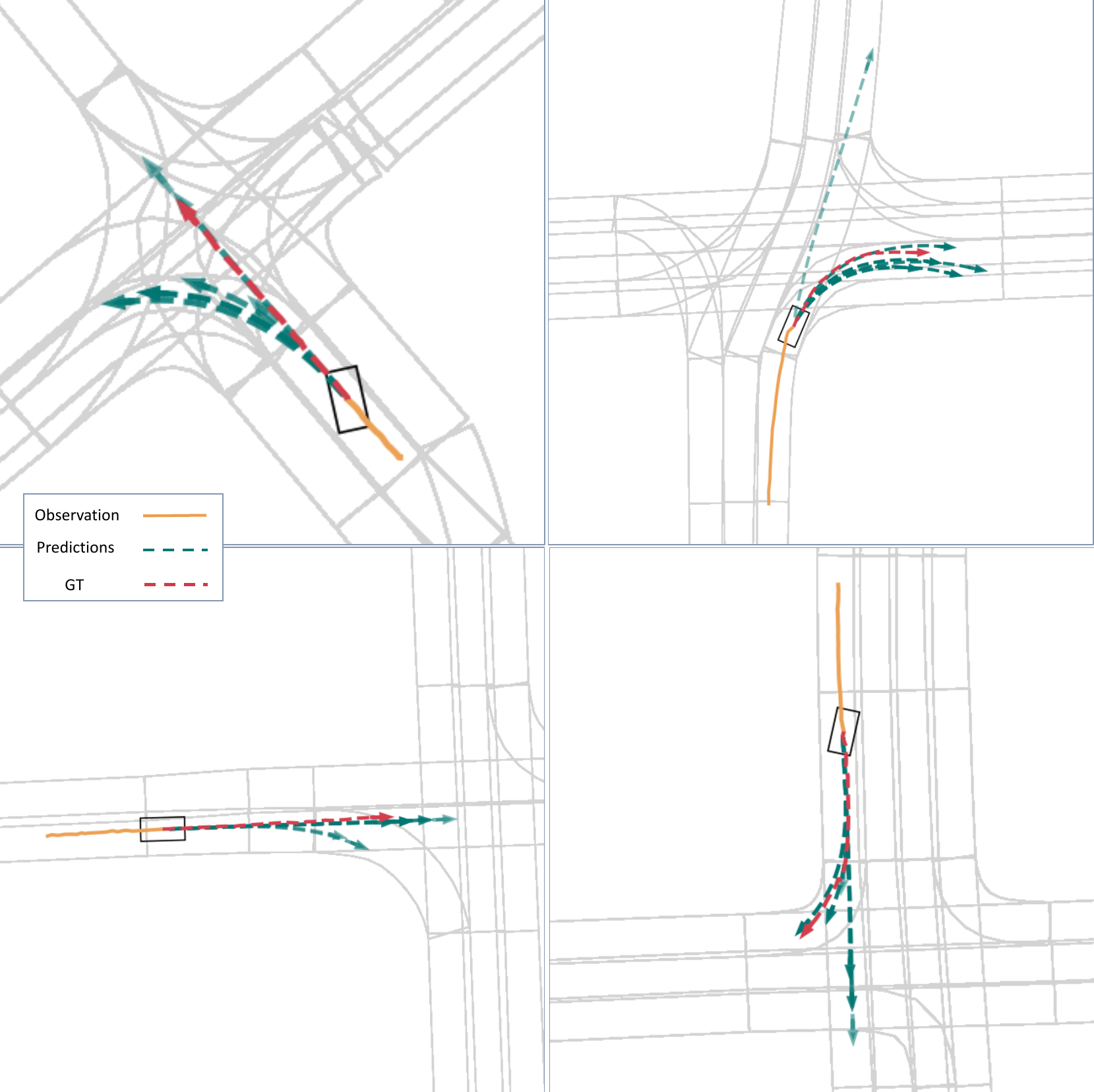}
	\end{center}
		\vspace{-0.2cm}
	\caption{Qualitative examples of the proposed method, DESTINE, on the Argoverse validation set.}
	\label{fig:qual}
\end{figure}

We present additional qualitative examples in Figure \ref{fig:qual}  to better show the performance of our proposed method under different conditions. Of particular interest, is the sample in top-left corner in which the intention of the vehicle changes towards the end of the observation period as shown by the orientation of the vehicle. In such a scenario, the model has a high-confidence of turning while  captures the possibility that the vehicle might move forward. In scenarios with turns in them, one can also notice the changes in the distribution of the modes. Often when the vehicle is closer to performing an action, e.g. turning point, the modes are more concentrated (as shown in top-right image) and the further the vehicle is, the modes are more apart to capture all possibilities (as shown in the examples in the second row).

{\small
	\bibliographystyle{IEEEtranS}
	\bibliography{references}
}

\end{document}